\documentclass[sigconf]{acmart}
\def\BibTeX{{\rm B\kern-.05em{\sc i\kern-.025em b}\kern-.08emT\kern-.1667em\lower.7ex\hbox{E}\kern-.125emX}}
    
\usepackage{nicefrac}
\usepackage{siunitx}
\usepackage{array,framed}
\usepackage{booktabs}
\usepackage{
  color,
  float,
  epsfig,
  wrapfig,
  graphics,
  graphicx,
  subcaption
}

\usepackage{textcomp,amssymb}
\usepackage{setspace}
\usepackage{latexsym,fancyhdr,url}
\usepackage{enumerate}
\usepackage{algorithm2e}
\usepackage{algpseudocode}
\usepackage{graphics}
\usepackage{xparse} 
\usepackage{xspace}
\usepackage{multirow}
\usepackage{csvsimple}
\usepackage{balance}

\usepackage{wrapfig}
\input{insbox}
\usepackage[symbol]{footmisc}

\usepackage{
  tikz,
  pgfplots,
  pgfplotstable
}
\usepackage{hyperref}

\usetikzlibrary{
  shapes.geometric,
  arrows,
  external,
  pgfplots.groupplots,
  matrix
}

\pgfplotsset{compat=1.9}

\usepackage{mathtools,}

\DeclareMathAlphabet{\mathcal}{OMS}{cmsy}{m}{n}

\setcopyright{none}
\renewcommand\footnotetextcopyrightpermission[1]{} 

\DeclareGraphicsExtensions{%
    .png,.PNG,%
    .pdf,.PDF,%
    .jpg,.mps,.jpeg,.jbig2,.jb2,.JPG,.JPEG,.JBIG2,.JB2}

\usepackage{xparse}
\newcommand{\bnm}{\begin{newmath}}
\newcommand{\enm}{\end{newmath}}

\newcommand{\bea}{\begin{eqnarray*}}%
\newcommand{\eea}{\end{eqnarray*}}%

\newcommand{\bne}{\begin{newequation}}
\newcommand{\ene}{\end{newequation}}

\newcommand{\bal}{\begin{newalign}}
\newcommand{\eal}{\end{newalign}}

\newenvironment{newalign}{\begin{align}%
\setlength{\abovedisplayskip}{4pt}%
\setlength{\belowdisplayskip}{4pt}%
\setlength{\abovedisplayshortskip}{6pt}%
\setlength{\belowdisplayshortskip}{6pt} }{\end{align}}

\newenvironment{newmath}{\begin{displaymath}%
\setlength{\abovedisplayskip}{4pt}%
\setlength{\belowdisplayskip}{4pt}%
\setlength{\abovedisplayshortskip}{6pt}%
\setlength{\belowdisplayshortskip}{6pt} }{\end{displaymath}}

\newenvironment{newequation}{\begin{equation}%
\setlength{\abovedisplayskip}{4pt}%
\setlength{\belowdisplayskip}{4pt}%
\setlength{\abovedisplayshortskip}{6pt}%
\setlength{\belowdisplayshortskip}{6pt} }{\end{equation}}

\newcounter{ctr}

\newcounter{mytable}
\def\mytable{\begin{centering}\refstepcounter{mytable}}
\def\endmytable{\end{centering}}

\newcounter{myfig}
\def\myfig{\begin{centering}\refstepcounter{myfig}}
\def\endmyfig{\end{centering}}

\newlength{\saveparindent}
\newlength{\saveparskip}
\newcommand{\E}{{\rm I\kern-.3em E}}

\renewcommand{\eqref}[1]{\mbox{Equation~(\ref{#1})}}

\def \part {part}

\renewcommand{\paragraph}[1]{\vspace*{6pt}\noindent\textbf{#1}\;}

\def \blackslug{\hbox{\hskip 1pt \vrule width 4pt height 8pt
    depth 1.5pt \hskip 1pt}}
\def \qed{\quad\blackslug\lower 8.5pt\null\par}

\newcounter{mynote}[section]

\newcommand\ignore[1]{}

\newcounter{rcnote}[section]

\newcounter{mrnote}[section]

\newcounter{fknote}[section]

\newcounter{anote}[section]

\DeclareMathSymbol{\mlq}{\mathord}{operators}{``}
\DeclareMathSymbol{\mrq}{\mathord}{operators}{`'}

\newcommand{\rhf}[2]{R_{f, \gamma}}

\DeclareDocumentCommand{\edist}{o o}{
  \ensuremath{
    \IfNoValueTF{#1}{{d}}{{\sf d}(#1,#2)}
  }
}

\newcommand{\olrk}[1]{\ifx\nursymbol#1\else\!\!\mskip4.5mu plus 0.5mu\left(\mskip0.5mu plus0.5mu #1\mskip1.5mu plus0.5mu \right)\fi}

\NewDocumentCommand{\indseq}{ O{1} O{r} }{{#1}\ldots {#2}}

\begin{document}
\fancyhead{}
\def\thetitle{AI Mobile Application for Archaeological Dating of Bronze Dings} 
\title{\thetitle}

\author{\small{Chuntao Li\textsuperscript{\textasteriskcentered}, Ruihua Qi\textsuperscript{\textasteriskcentered}, Chuan Tang\textsuperscript{\textasteriskcentered}, Jiafu Wei\textsuperscript{\textasteriskcentered}, Xi Yang\textsuperscript{\textasteriskcentered}, Qian Zhang\textsuperscript{\textasteriskcentered}, Rixin Zhou}}
\authornote{Equal contribution. Listing order is alphabetical.}

\affiliation{\small{Jilin University}}

\date{}

\begin{abstract}
\InsertBoxR{9}{\includegraphics[width=0.3\linewidth]{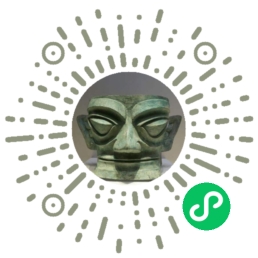}}
We develop an AI application for archaeological dating of bronze Dings. A classification model is employed to predict the period of the input Ding, and a detection model is used to show the feature parts for making a decision of archaeological dating. To train the two deep learning models, we collected a large number of Ding images from published materials, and annotated the period and the feature parts on each image by archaeological experts. Furthermore, we design a user system and deploy our pre-trained models based on the platform of WeChat Mini Program for ease of use. Only need a smartphone installed WeChat APP, users can easily know the result of intelligent archaeological dating, the feature parts, and other reference artifacts, by taking a photo of a bronze Ding. To use our application, please scan this QR code by WeChat.

\end{abstract}

\maketitle

\section{Introduction}

In archaeology, typology exploits the shape and decoration of an artifact to deduce its period. Dating artifacts depends on the long-term training and accumulation of experts, and different judgments can be made based on personal knowledge frequently. In this application, focusing on Chinese bronze Dings, we explore an automatic archaeology dating solution by using deep learning models.

Dings are cauldrons that are used for cooking, storage, and ritual offerings to the gods or to ancestors in ancient China, they are one of the most important species used in Chinese ritual bronzes~\cite{wiki:Ding_(vessel)}. The dating of Dings contributes to studying ancient Chinese history. For providing more useful information, we also mark out the important parts and give reference artifacts for inference besides the dating result. Considering the feasibility in practise, we select image data rather than 3D surface models as inputs, although the latter can capture better shape and texture information for a 3D object. 

To sufficiently utilize state-of-the-art deep learning techniques, we collect a large number of images of Dings, and re-argue the period of each artifact discussing with Bronze experts. For easy use to both experts and common people, we deploy a system based on the WeChat Mini program (Version 1.0) ~\cite{mini}, a Ding can be inferred conveniently by taking a photo with a smartphone. Our application is developed for study and education, not commercial.

\section{Application}

\subsection{Deployment}

\begin{figure}[t]
  \includegraphics[width=0.42\linewidth]{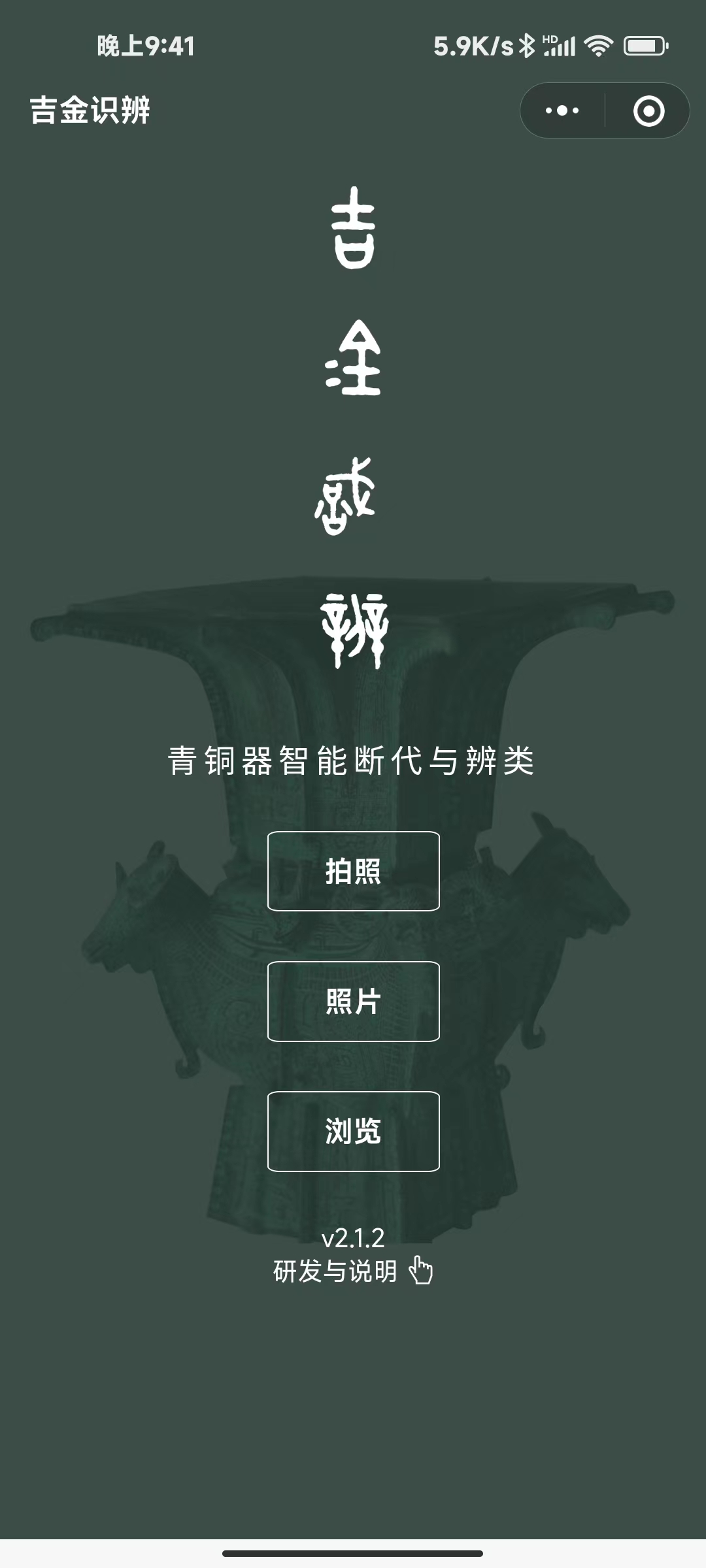}
  \includegraphics[width=0.42\linewidth]{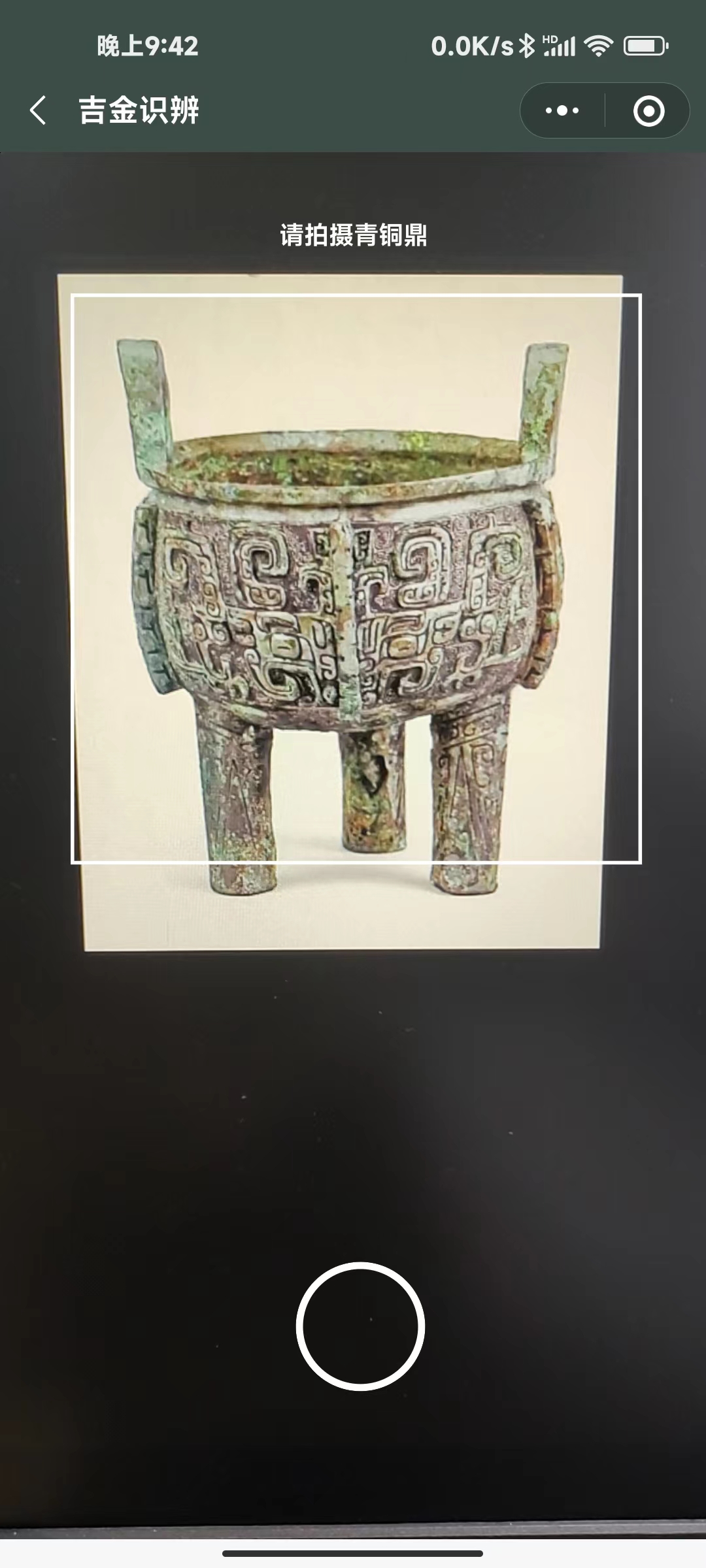}
  \includegraphics[width=0.42\linewidth]{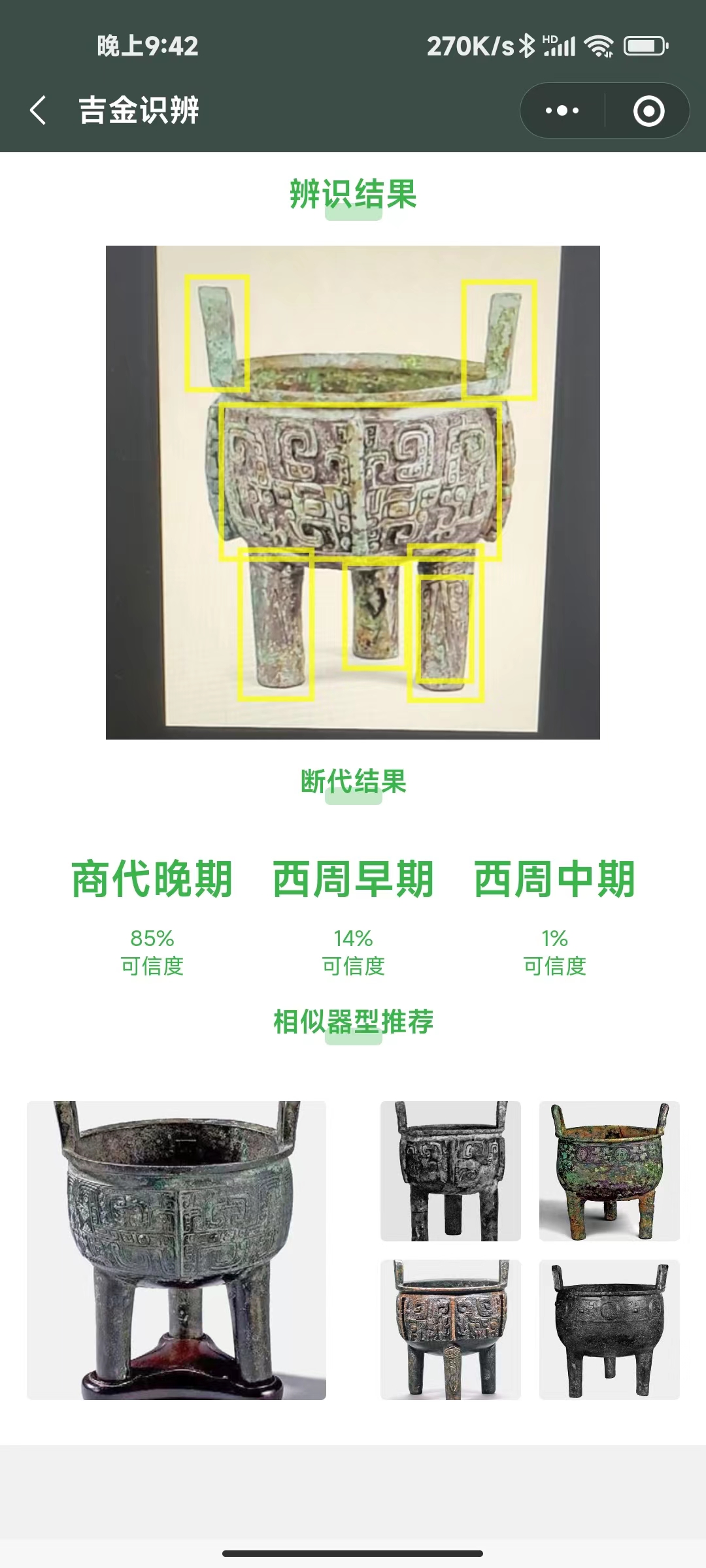}
  \includegraphics[width=0.42\linewidth]{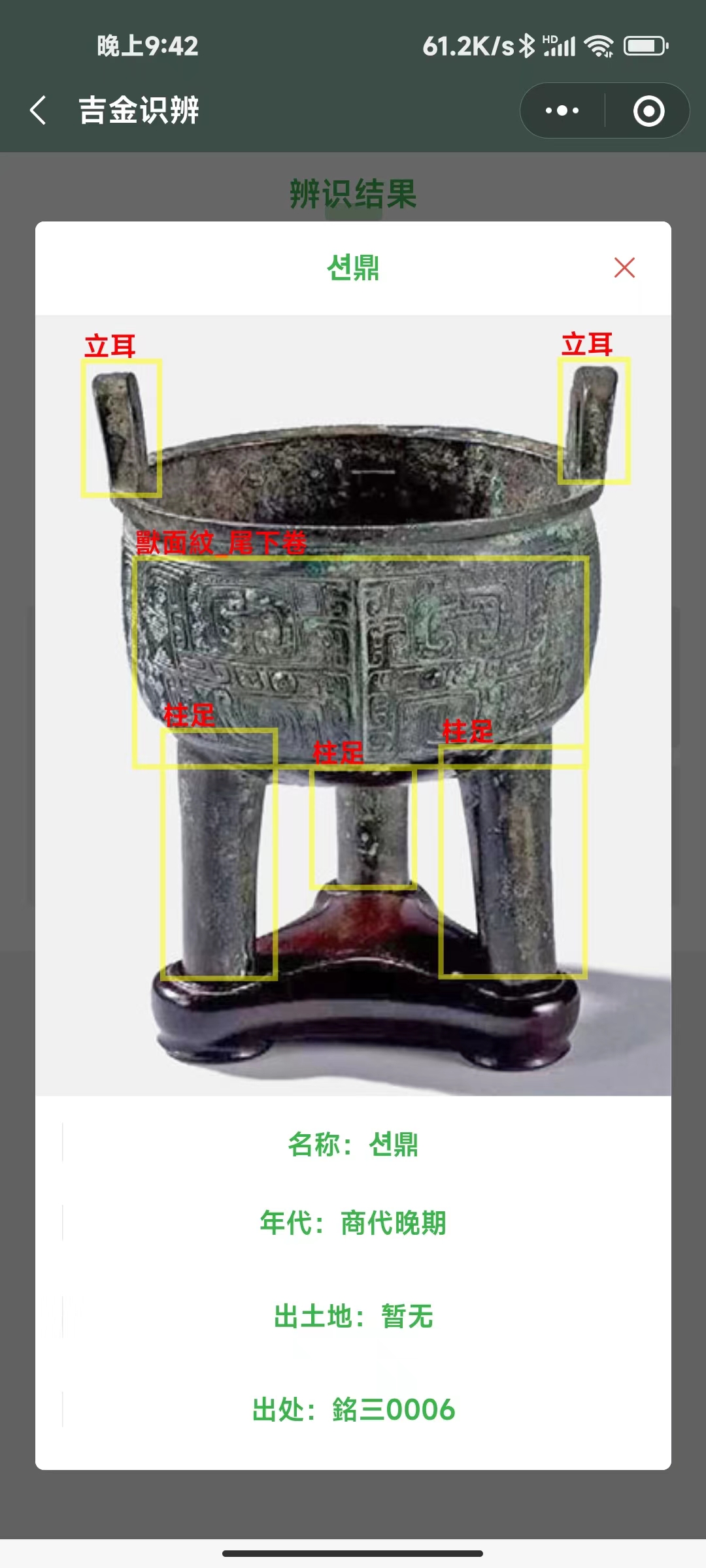}
  \caption{Interface of our application (in Chinese). The top left figure shows the main page, user can upload a photo by pressing the middle buttons 
  , as shown in the top right figure. And then, the dating results are shown as the bottom left figure. The inferred period is shown in the middle, the feature parts are marked by yellow boxes in the top figure, and the reference artifacts are shown in the bottom. Pressing a reference artifact, its information is shown as the bottom right figure.}
  \label{fig:interface}
\end{figure}

We design a user system based on Wechat Mini Program~\cite{mini} as shown in Figure~\ref{fig:interface}. And we train image-based deep learning models for predication and deploy them on our local sever. By pressing the middle buttons in main page of our application, user can take a photo or select a photo from smartphone. Then, this image is sent to our sever and fed into pre-trained models. After prediction, the results are shown in the application including inferred period, feature parts, and reference artifacts.

\begin{table*}[t]
\centering
\caption{Data and results. The image numbers are the complete dataset we collected, and the accuracies are tested on our testset.}
\label{tab:periods}
\begin{tabular}{l|c|ccccccccccc}
\hline
\multirow{2}{*}{Period} & \multirow{2}{*}{Total} & \multicolumn{2}{c}{Shang}                   & \multicolumn{3}{c}{Western Zhou}                                   & \multicolumn{3}{c}{Spring and Autumn}                              & \multicolumn{3}{c}{Warring States}                                 \\ 
                     &                        & \multicolumn{1}{l}{Early} & \multicolumn{1}{l}{Late} & \multicolumn{1}{l}{Early} & \multicolumn{1}{l}{Mid} & \multicolumn{1}{l}{Late} & \multicolumn{1}{l}{Early} & \multicolumn{1}{l}{Mid} & \multicolumn{1}{l}{Late} & \multicolumn{1}{l}{Early} & \multicolumn{1}{l}{Mid} & \multicolumn{1}{l}{Late} \\ \hline
Number               &         $\approx$ 4,000    &       2.3\%               &         27.7\%             &        24.2\%              &          12.1\%            &          7.6\%            &        8.0\%              &           4.8\%           &           5.7\%           &          2.1\%            &            1.7\%          &         3.8\%             \\
Accuracy             &        81.41\%                &         100.00\%             &        95.83\%              &          87.50\%            &         89.74\%             &          84.84\%            &        80.76\%              &          54.16\%            &          73.91\%            &       87.5\%               &           42.85\%           &         85.71\%            \\ \hline
\end{tabular}
\end{table*}

\begin{figure*}[t]
  \includegraphics[width=0.95\linewidth]{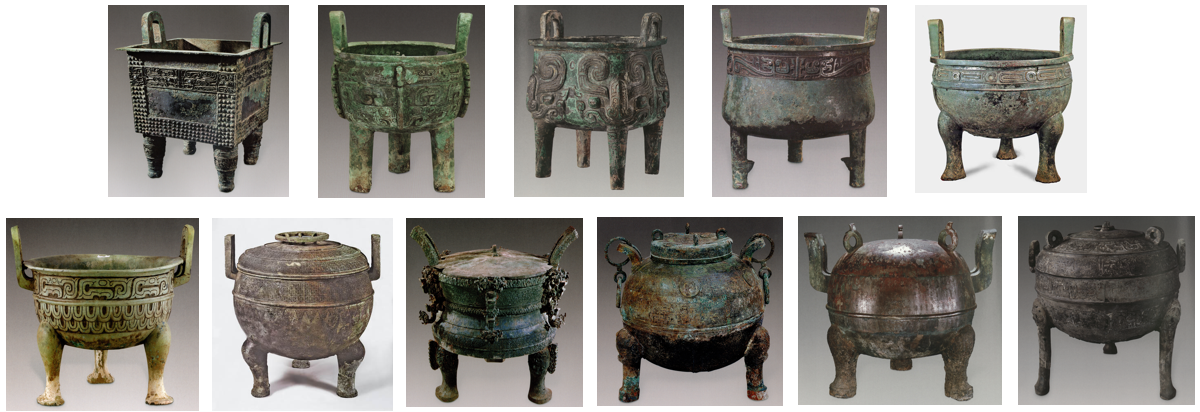}
  \caption{Examples of bronzes from 11 periods in the order as Table~\ref{tab:periods}. }
  \label{fig:examples}
\end{figure*}

\subsection{Data}

We collect about 4 thousands images of Dings from published archaeology books and websites, examples for each age are shown in the Figure~\ref{fig:examples}. These Dings are divided into 11 periods and the details are shown in Table~\ref{tab:periods}. Each image is annotated by bronze experts, including period, shape, literature, excavation, museum. The bounding boxes of the feature parts are also marked on each image by using a annotation tool of object detection, such as handles, legs, and decoration. 

For pre-processing, we apply data augmentation to the collected images to improve the prediction results, including background removal, gray-scale and feature line extraction.

\vspace*{-10pt}
\subsection{Inference}

With comparison, we employ ConvNeXt~\cite{convnext} to learn the features of shape and decoration on Ding and predict their age, and use SparseR-CNN~\cite{sparsercnn} to detect the feature parts of bronzes.

To test our pre-trained models, experts collect 300 images consisting of a uniform distribution of each period as a testset. The results of dating accuracy are shown in Table~\ref{tab:periods}. Because the number and dating difficulty of each period is quite unbalance, the difference among prediction results is significant. However, the infer is satisfactory for not rare ones.

In practice, because many Dings are ambiguous between adjacent ages even in archaeology, we show top four dating predictions on our application instead of only the top one. And we set the prediction result is "other stuffs" if the top one predicted probability is smaller than 5\%. For the reference artifacts, we select the top five by computing and sorting the similarity of the feature vectors from the classification encoder.






\section{Discussion}

We found that the predication results with separated classification and detection models are better than a single model. Therefore, we extract different features for period prediction and feature parts prediction.

The shapes and decorations are similar and complex whether in different periods or within the same period, even a human requires long-term learning to make decision, and some objects are hard to into an period only rely on its shape even experts. However, SOTA networks performed satisfactory accuracy with only inputting entire images. And the error cased are reasonable. 

The view angle of the input image will affect the dating result, since our collected images of Dings are almost taken from the front. However, the impact is not significant if the image is not taken from above or below.

The current application focuses on the Bronze Dings, in the further work, we will study the whole category of bronzes.

\section{Summary}

In summary, we developed an AI mobile application for the dating of bronze Dings by collecting images and deep learning network. Welcome to scan the QR code.

\bibliographystyle{unsrt}

\bibliographystyle{ACM-Reference-Format}
\bibliography{ref}

\appendix

\end{document}